\documentclass[11pt,a4paper]{article}
\usepackage[hyperref]{acl2019}
\usepackage{times}
\usepackage{latexsym}
\usepackage{url}
\usepackage{subfiles}
\usepackage{todonotes}
\usepackage{wrapfig}
\usepackage{subfig}
\usepackage{float}
\usepackage{caption}
\usepackage{tikz}
\usetikzlibrary{decorations.pathreplacing}
\usepackage{graphicx}

\usepackage[utf8]{inputenc} 
\usepackage[T1]{fontenc}    
\usepackage{hyperref}       
\usepackage{booktabs}       
\usepackage{amsfonts}       
\usepackage{nicefrac}       
\usepackage{microtype}      

\aclfinalcopy 
\def\aclpaperid{1891}

\title{Multimodal Abstractive Summarization for How2 Videos}

\newcommand*{\affaddr}[1]{#1} 
\newcommand*{\affmark}[1][*]{\textsuperscript{#1}}
\newcommand*{\email}[1]{\texttt{#1}}

\author{}

\author{Shruti Palaskar\affmark[1] \quad Jindřich Libovický\affmark[2] \quad Spandana Gella\affmark[3]\thanks{*Work done while SG was at University of Edinburgh} 
\quad Florian Metze\affmark[1]\\
\affaddr{\affmark[1]School of Computer Science, Carnegie Mellon University}\\
\affaddr{\affmark[2]Faculty of Mathematics and Physics, Charles University}\\
\affaddr{\affmark[3] Amazon AI}\\
\email{spalaska@cs.cmu.edu, libovicky@ufal.mff.cuni.cz}\\
\email{sgella@amazon.com, fmetze@cs.cmu.edu} 
}

\date{}

\begin{document}
\maketitle
\begin{abstract}
  In this paper, we study abstractive summarization for open-domain videos.
Unlike the traditional text news summarization, the goal is less to ``compress'' text information but rather to provide a fluent textual summary of information that has been collected and fused from different source modalities, in our case video and audio transcripts (or text).
We show how a multi-source sequence-to-sequence model with hierarchical attention can integrate information from different modalities into a coherent output, compare various models trained with different modalities and present pilot experiments on the \emph{How2 corpus} of instructional videos. We also propose a new evaluation metric (Content F1) 
for abstractive summarization task that measures semantic adequacy rather than fluency of the summaries, which is covered by metrics like ROUGE and BLEU.

\end{abstract}

\section{Introduction}
\label{sec:intro}
In recent years, with the growing popularity of video sharing platforms, 
there has been a steep rise in the number of user-generated instructional 
videos shared online. With the abundance of videos online, there has been an increase in 
demand for efficient ways to search and retrieve relevant videos \cite{song2011multiple,wang2012event,otani2016learning,TorabiTS16}. 
Many cross-modal search applications rely on text associated with the video such as description or
title to find relevant content. However, often videos do not have text meta-data associated with them or the existing ones do not provide clear information of the video content and 
fail to capture subtle differences between related videos \cite{wang2012event}.  
We address this by aiming to generate a short text summary of the video that describes the most salient content of the video.
Our work benefits users through better contextual information and user experience, and video sharing platforms with increased user engagement by retrieving or suggesting 
relevant videos to users and capturing their attention.

\begin{figure*}[t]
  \centering
  \includegraphics[ trim={0 5cm 0 0}, clip, width=\linewidth]{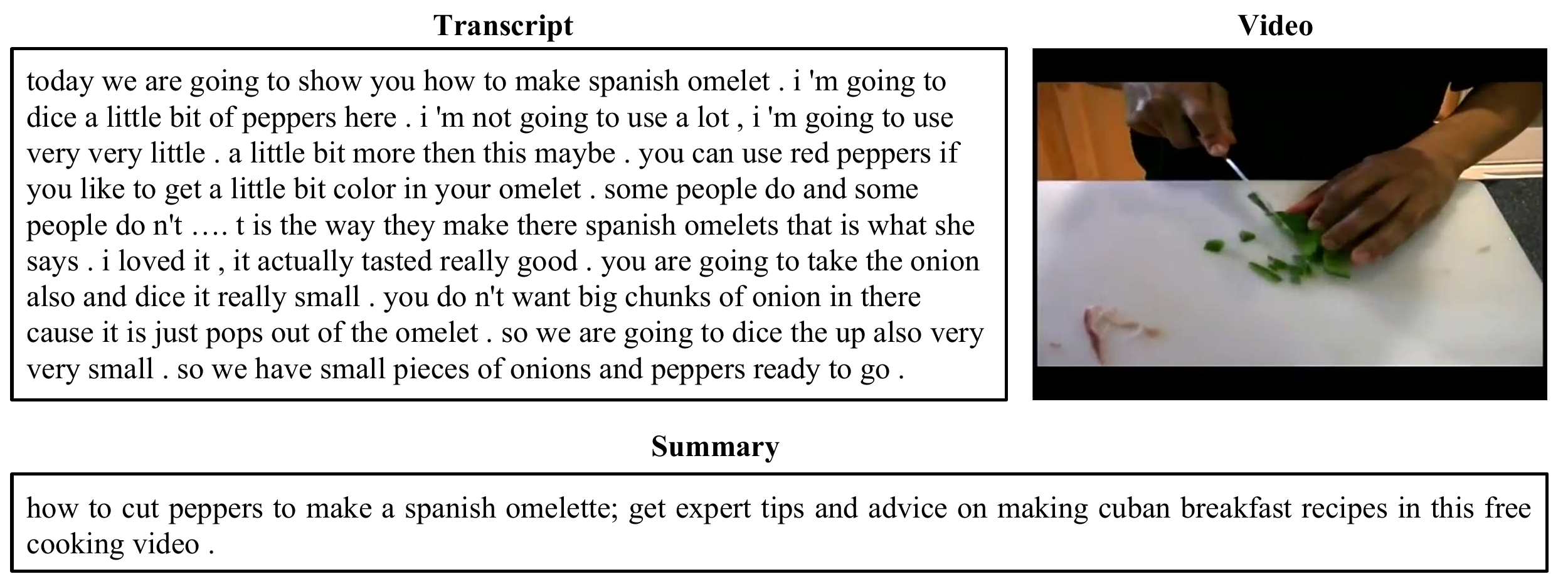}
  \caption{How2 dataset example with different modalities. ``Cuban breakfast'' and ``free cooking video'' is not mentioned in the transcript, and has to be derived from other sources.}
  \label{fig:how2data}
\end{figure*}

Summarization is a task of producing a shorter version of the content in the document while preserving its information and has been studied for both textual documents (automatic text summarization) and visual documents such as images and videos (video summarization). Automatic text summarization is a widely studied topic in natural language processing \cite{luhn1958automatic,kupiec1995trainable,mani1999advances}; given a text document the task is to generate a textual summary for applications that can assist users to understand large documents.
Most of the work on text summarization has focused on single-document 
summarization for domains such as news \cite{rush2015neural,cnndm_nallapathi,see2017get,narayan2018ranking} and some on multi-document summarization \cite{goldstein2000multi,lin2002single,woodsend2012multiple,cao2015ranking,YasunagaZMPSR17}.

Video summarization is the task of producing a compact version of the video (visual summary) by encapsulating the most informative parts 
\cite{money2008video,lu2013story,gygli2014creating,song2015tvsum,sah2017semantic}.
Multimodal summarization is the combination of textual and visual modalities by summarizing a video document with a text summary that summarizes the content of the video.
Multimodal summarization is a more recent challenge with no benchmarking datasets yet. \citet{li2017multi} collected a multimodal corpus of 500 English news videos and articles paired with manually annotated summaries. The dataset is small-scale and has news articles with audio, video, and text summaries, but there are no human annotated audio-transcripts.

Related tasks include image or video captioning and  description generation, video story generation, procedure learning from instructional videos and title generation which focus on events or activities in the video and generating descriptions at various levels of granularity from single sentence to multiple sentences \cite{DaXuDoCVPR2013,RegneriRWTSP13,RohrbachRQFPS14,zeng2016generation,ZhouXC18,zhang2018product,gella2018dataset}. A closely related task to ours is video title generation where the task is to describe the most salient event in the video in a compact title that is aimed at capturing users attention \cite{zeng2016generation}. \citet{ZhouXC18} present the YouCookII dataset containing instructional videos, specifically cooking recipes, with temporally localized annotations for the procedure which could be viewed as a summarization task as well although localized with time alignments between video segments and procedures.  

In this work, we study multimodal summarization with various methods to summarize the intent of \emph{open-domain instructional videos} stating the exclusive and unique features of the video, irrespective of modality. We study this task in detail using the new How2 dataset \citep{how2} which contains human annotated video summaries for a varied range of topics. Our models generate natural language descriptions for video content using the transcriptions (both user-generated and output of automatic speech recognition systems) as well as visual features extracted from the video. We also introduce a new evaluation metric (Content F1) that suits this task and present detailed results to understand the task better.

\section{Multimodal Abstractive Summarization}
\label{sec:dataset}
The How2 dataset \citep{how2} contains about 2,000 hours of short instructional
videos, spanning different domains such as cooking, sports, indoor/outdoor
activities, music, etc. Each video is accompanied by a human-generated transcript 
and a 2 to 3 sentence
summary is available for every video written to generate interest in a potential viewer. 

The example in Figure~\ref{fig:how2data} shows the transcript describes instructions in detail, while the summary is a high-level overview of the entire video, mentioning that the peppers are being ``cut'', and that this is a ``Cuban breakfast recipe'', which is not mentioned in the transcript. We observe that text and vision modalities both contain complementary information, thereby when fused, helps in generating richer and more fluent summaries. Additionally, we can also leverage the speech modality by using the output of a speech recognizer as input to a summarization model instead of a human-annotated transcript. 

The How2 corpus contains 73,993 videos for training, 
2,965 for validation and 2,156 for testing. The average length of transcripts is 291 words and of summaries is 33 words. A more general comparison of the How2 dataset for summarization as compared with certain common datasets is given in \citep{how2}. 

\paragraph{Video-based Summarization.} We represent videos by features extracted from a pre-trained action recognition model: a ResNeXt-101 3D Convolutional Neural Network \citep{hara3dcnns}
trained to recognize 400 different human actions in the Kinetics dataset \citep{kay2017kinetics}. These features are 2048 dimensional, extracted for every 16 non-overlapping
frames in the video. This results in a sequence of feature vectors per video rather than a single/global one. We use these sequential features in our models described in Section \ref{sec:methods}.  2048-dimensional feature vector representing all text a single video.

\paragraph{Speech-based Summarization.} We leverage the speech modality by using the outputs from a pre-trained speech recognizer that is trained with other data, as inputs to a text summarization model. We use the state-of-the-art models for distant-microphone conversational speech recognition, ASpIRE \citep{peddinti2015jhu} and EESEN \citep{miao2015eesen,le2018aclew}. The word error rate of these models on the How2 test data is 35.4\%. This high error mostly stems from normalization issues in the data.
For example, recognizing and labeling ``20'' as ``twenty'' etc. Handling these effectively will reduce the word error rates significantly. We accept these as is for this task.

\paragraph{Transfer Learning.} Our parallel work \citet{sanabria2019cmu} demonstrates the use of summarization models trained in this paper for a transfer learning based summarization task  on the Charades dataset \citep{sigurdsson2016hollywood} that has audio, video, and text (summary, caption and question-answer pairs) modalities similar to the How2 dataset. \citet{sanabria2019cmu} observe that pre-training and transfer learning with the How2 dataset led to significant improvements in unimodal and multimodal adaptation tasks on the Charades dataset.

\section{Summarization Models}
\label{sec:methods}
\begin{figure}[t]
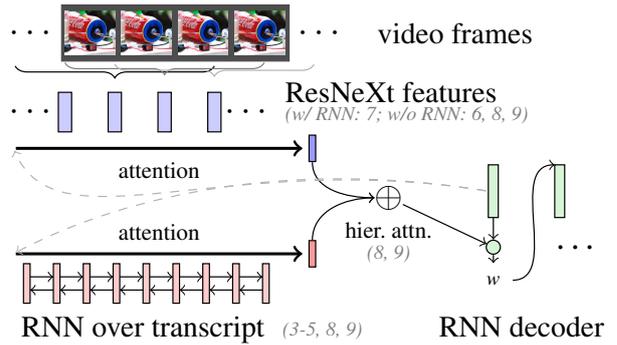

\scalebox{0.0}{\input{final/model}}%
\begin{center}\hspace*{-30pt}\scalebox{1.13}{\input{final/model}} \end{center}
\caption{Building blocks of the sequence-to-sequence models, gray numbers in
brackets indicate which components are utilized in which
experiments.}\label{fig:architectures}
\end{figure}

We study various summarization models. First, we use a 
Recurrent Neural Network (RNN) Sequence-to-Sequence (S2S) model \cite{Sutskever2014} consisting of 
an encoder RNN to encode (text or video features) with the attention mechanism \citep{bahdanau2014neural} and a decoder RNN to generate
summaries. Our second model is a Pointer-Generator (PG) model \cite{vinyals2015pointer,GulcehreANZB16} that has shown strong performance for abstractive summarization \citep{cnndm_nallapathi,see2017get}. As our third model, we use hierarchical attention approach of \citeauthor{libovicky2017attention} \citeyear{libovicky2017attention}  originally proposed for multimodal machine translation to combine textual and visual modalities to generate text. The model first computes the context vector independently for each of the input modalities (text and video). In the next step, the context vectors are treated as states of 
another encoder, and a new vector is computed. When using a sequence of action features 
instead of a single averaged vector for a video, the RNN layer helps capture context. In Figure~\ref{fig:architectures} we present the building block of our models.

\section{Evaluation}

\begin{table*}[t]
  \centering
  \scalebox{.95}{
  \begin{tabular}{clcc}
    \toprule
    Model No. & Description 					& ROUGE-L & Content F1 \\
    \midrule
    1     	& Random Baseline using Language Model 	& 27.5	& 8.3      \\
    2a 		& Rule-based Extractive summary 		& 16.4	& 18.8     \\
    2b      & Next-neighbor Summary                 & 31.8  & 17.9     \\
    \midrule
    3		& Using Extracted Sentence from 2a only (Text-only)	& 46.4	& 36.0 \\
    4		& First 200 tokens	(Text-only)					    & 40.3	& 27.5	\\
    5a		& S2S Complete Transcript (Text-only, 650 tokens)		& \textbf{53.9}	& \textbf{47.4}	\\
    5b		& PG Complete Transcript (Text-only)		& 50.2	& 42.0	\\
    5c		& ASR output Complete Transcript (Text-only)		& 46.1	& 34.7	\\
    \midrule
    6		& Action Features only (Video)	&	38.5	&	24.8	\\
    7		& Action Features + RNN (Video)	&	\textbf{46.3}	&   \textbf{34.9}	\\
    \midrule
    8		& Ground-truth transcript + Action with Hierarchical Attn		& \textbf{54.9}	& \textbf{48.9} \\
    9		& ASR output + Action with Hierarchical Attn		& 46.3	& 34.7 \\
    \bottomrule
  \end{tabular}}
  \caption{ROUGE-L and Content F1 for different summarization models: random baseline (1), rule-based extracted summary (2a), nearest neighbor summary (2b), different text-only (3,4,5a), pointer-generator (5b), ASR output transcript (5c), video-only (6-7) and text-and-video models (8-9).}
  \label{tab:results}
\end{table*}

We evaluate the summaries using the standard metric for
abstractive summarization ROUGE-L \citep{rouge}
that measures the longest common sequence between the reference and the generated summary. Additionally, we introduce the Content F1 metric that fits the template-like structure of the summaries. We analyze the most frequently occurring words in the transcription and 
summary. 
The words in transcript reflect the conversational and spontaneous speech while the words in the summaries reflect their descriptive nature. For examples, see Table~\ref{tab:freq} in Appendix \ref{ssec:supp_freq}. 

\paragraph{Content F1.} This metric is the F1 score of the content words in the
summaries based over a monolingual alignment, similar to metrics used to evaluate quality of monolingual alignment \citep{sultan2014back}.
We use the METEOR toolkit
\citep{banerjee2005meteor,denkowski2014meteor} to obtain the alignment.
Then, we remove function words and task-specific stop words that appear in most
of the summaries (see Appendix \ref{ssec:supp_freq}) from the reference and the hypothesis. The stop words are
easy to predict and thus increase the ROUGE score. We treat remaining content words from the reference and
the hypothesis as two bags of words and compute the F1 score over the alignment. 
Note that the score ignores the fluency of output.

\paragraph{Human Evaluation.} In addition to automatic evaluation, we perform a human evaluation to understand the outputs of this task better. Following the abstractive summarization human annotation work of  \citet{grusky2018newsroom}, we ask our annotators to label the generated output on a scale of $1-5$ on informativeness, relevance, coherence, and fluency. We perform this on randomly sampled 500 videos from the test set.

We evaluate three models: two unimodal (text-only (5a), video-only (7)) and one multimodal (text-and-video (8)). Three workers annotated each video on Amazon Mechanical Turk. More details about human evaluation are in the Appendix \ref{ssec:supp_human}.

\section{Experiments and Results}
\label{sec:results_discussion}
\begin{table}[t]
  \centering
  \scalebox{.95}{
  \begin{tabular}{lcccc}
    \toprule
    Model (No.)		& INF & REL & COH & FLU \\
    \midrule
    Text-only (5a)		& 3.86	& \bf{3.78}	& 3.78	& 3.92\\
    Video-only (7)		& 3.58	& 3.30  & 3.71	& 3.80\\
    Text-and-Video (8)	& 
    \bf{3.89}	& 3.74	& \bf{3.85}	& \bf{3.94}\\
    \bottomrule
    \end{tabular}}
    \caption{Human evaluation scores on 4 different measures of  Informativeness (INF), Relevance (REL), Coherence (COH), Fluency (FLU). }
    \label{tab:human_eval}
\end{table}

As a baseline, we train an RNN language model \citep{sutskever2011generating}
on all the summaries and randomly sample tokens from it.
The output obtained is fluent in English leading to a high ROUGE
score, but the content is unrelated which leads to a low Content F1 score in
Table~\ref{tab:results}. As another baseline, we replace the target summary
with a rule-based extracted summary from the transcription itself. We used the
sentence containing words ``how to'' with predicates \emph{learn}, \emph{tell},
\emph{show}, \emph{discuss} or \emph{explain}, usually the second sentence in
the transcript. Our final baseline was a model trained with the summary of
the nearest neighbor of each video in the Latent Dirichlet Allocation (LDA; \citealp{blei2003latent}) based topic space as a target. This model achieves a
similar Content F1 score as the rule-based model which shows the similarity of
content and further demonstrates the utility of the Content F1 score.

We use the transcript (either ground-truth transcript or speech recognition output) and the video action features to train various models with different combinations of modalities. The text-only model performs best when using the complete
transcript in the input (650 tokens). This is in contrast to prior work with news-domain summarization \citep{cnndm_nallapathi}. We also observe that PG networks do not perform better than S2S models on this data which could be attributed to the abstractive nature of our summaries and also the lack of common $n$-gram overlap between input and output which is the important feature of PG networks. We also use the automatic transcriptions obtained from a pretrained automatic speech recognizer as input to the summarization model. This model achieves competitive performance with the video-only models (described below) but degrades noticeably than ground-truth transcription summarization model. This is as expected due to the large margin of ASR errors in distant-microphone open-domain speech recognition.

We trained two video-only models: the first one uses a
single mean-pooled feature vector representation for the entire video, while the second one applies a
single layer RNN over the vectors in time.

Note that using only the action features in input reaches
almost competitive ROUGE and Content F1 scores compared to the text-only model showing
the importance of both modalities in this task. Finally, the hierarchical
attention model that combines both modalities obtains the highest score.

In Table~\ref{tab:human_eval}, we report human evaluation scores on our best text-only, video-only and multimodal models. In three evaluation measures, the multimodal models with the hierarchical attention reach the best scores. Model hyperparameter settings, attention analysis and example outputs for the models described above are available in the Appendix.

\begin{figure}[t]
  \centering
  \includegraphics[width=\linewidth]{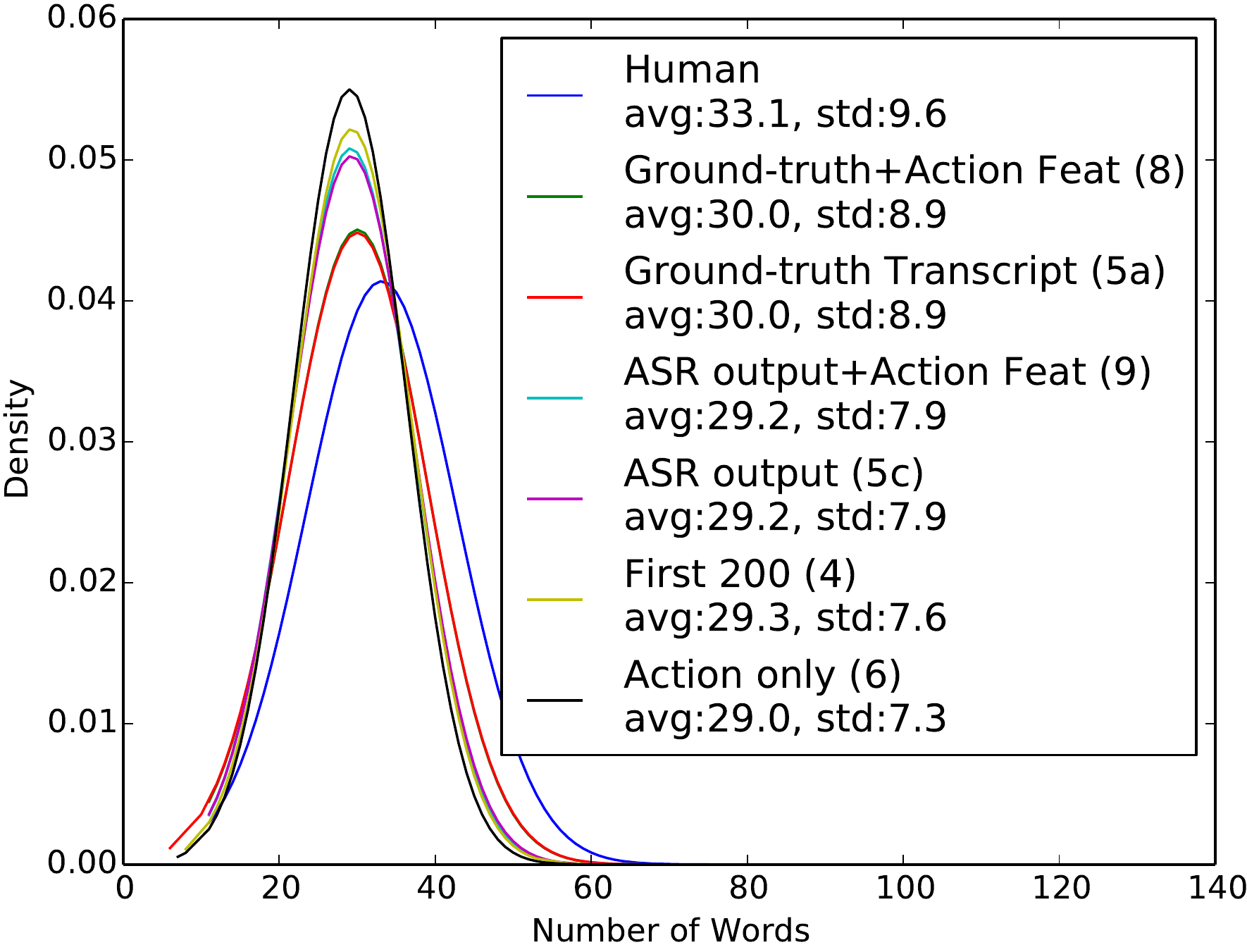}
  \caption{Word distribution in comparison with the human summaries for different unimodal and multimodal models. Density curves show the length distributions of human annotated and system produced summaries.}
  \label{fig:density}
\end{figure}

In Figure~\ref{fig:density}, we analyze the word distributions of different system generated summaries with the human annotated reference. The density curves show that most model outputs are shorter than human annotations with the action-only model (6) being the shortest as expected. Interestingly, the two different uni-modal and multimodal systems with ground-truth text and ASR output text features are very similar in length showing that the improvements in Rouge-L and Content-F1 scores stem from the difference in content rather than length. Example presented in Table \ref{tab:output_examples}  Section~\ref{ssec:supp_outputs} shows how the outputs vary.

\section{Conclusions}
\label{sec:conclusiom}
We present several baseline models for 
generating abstractive text summaries for the open-domain videos in How2 data. 
Our presented models include a video-only summarization model that performs competitively with a text-only model. In the future, we would like to extend this work to generate multi-document (multi-video) summaries and also build end-to-end models directly from audio in the video instead of text-based output from pretrained ASR. We define and show the quality of a new metric, Content F1, 
for evaluation of the video summaries that are designed as teasers or highlights 
for viewers, instead of a condensed version of the input like traditional 
text summaries.

\section*{Acknowledgements}

This work was mostly conducted at the 2018 Frederick 
Jelinek Memorial Summer Workshop on
Speech and Language Technologies,\footnote{https://www.clsp.jhu.edu/workshops/18-workshop/}
hosted and sponsored by Johns Hopkins University. Shruti Palaskar received funding from Facebook and Amazon grants. Jindřich Libovický received funding from the Czech Science Foundation, grant no. 19-26934X. This work used the Extreme Science and Engineering Discovery Environment (XSEDE) supported by NSF grant ACI-1548562 and the Bridges system supported by NSF award ACI-1445606, at the Pittsburgh Supercomputing Center.

\bibliography{refs}
\bibliographystyle{acl_natbib}

\appendix

\section{Appendix}
\subsection{Experimental Setup}
\label{sec:app-experment-setup}

In all our experiments, the text encoder consists of 2 bidirectional layers of
the encoder with 256 Gated Recurrent Units (GRU; \citealt{cho2014gru}) and 2
layers of the decoder with Conditional Gated Recurrent Units (CGRU;
\citealt{nematus}). We optimize the models with the Adam Optimizer
\citep{kingma2015adam} with learning rate $4\cdot10^{-4}$ halved after each
epoch when the validation performance does not increase for maximum 50 epochs.

We restrict the input
length to 600 tokens for all experiments except the best text-only model in the  
section Experiments and Results. We use vocabulary the 20,000 most
frequently occurring words which showed best results in our experiments, largely outperforming models using subword-based
vocabularies. We ran all experiments with the \texttt{nmtpytorch} toolkit \citep{nmtpy2017}.

\subsection{Frequent Words in Transcripts and Summaries}
\label{ssec:supp_freq}
Table \ref{tab:freq} shows the frequent words in transcripts (input) and summaries (output). The words in transcripts reflect conversational and spontaneous speech while words in the summary reflect their descriptive nature.

\begin{table}[t]
  \renewcommand{\thetable}{A1}
  \centering
  \begin{tabular}{p{1.3cm}p{5.5cm}}
    \toprule
    Set & Words 	 \\
    \midrule
    Transcript 	&  
    the, to, and, you, a, it, that, of, is, i, going, we, in, your, this, 's, so, on\\
    Summary		&  
    in, a, this, to, free, the, video, and, learn, from, on, with, how, tips, for, of, expert, an\\
    \bottomrule
  \end{tabular}
  \caption{Most frequently occurring words in Transcript and Summaries.}
  \label{tab:freq}
\end{table}

\begin{table*}[t]
  \centering
  \renewcommand{\thetable}{A2}
\begin{tabular}{cp{3cm}ccp{8.5cm}}
    \toprule
    No. & Model & R-L & C-F1 & Output \\
    \midrule
    - & Reference 	& -	& - & \small watch and learn how to tie thread to a hook to help with fly tying as explained by out expert in this free how - to video on fly tying tips and techniques . \\ [0.2cm]
    8 & Ground-truth text + Action Feat. & 54.9 & 48.9 & \small learn from our expert how to attach thread to fly fishing for fly fishing in this free how - to video on fly tying tips and techniques . \\ [0.2cm] 
    5a & Text-only (Ground-truth)	&  53.9 & 47.4 & \small learn from our expert how to tie a thread for fly fishing in this free how - to video on fly tying tips and techniques .	\\ [0.2cm]
    9 & ASR output + Action Feat. & 46.3 & 34.7 & \small learn how to tie a fly knot for fly fishing in this free how-to video on fly tying tips and techniques . \\ [0.2cm]
    5c & ASR output & 46.1 & 34.7 & \small learn tips and techniques for fly fishing in this free fishing video on techniques for and making fly fishing nymphs . \\ [0.2cm]
  	7 & Action Features + RNN	& 46.3 & 34.9 & \small learn about the equipment needed for fly tying , as well as other fly fishing tips from our expert in this free how - to video on fly tying tips and techniques .	\\ [0.2cm]
    6 & Action Features only	& 38.5 & 24.8 	& \small learn from our expert how to do a double half hitch knot in this free video clip about how to use fly fishing .	\\ [0.2cm]
   2b & Next Neighbor & 31.8 & 17.9 & \small use a sheep shank knot to shorten a long piece of rope . learn how to tie sheep shank knots for shortening rope in this free knot tying video from an eagle scout . \\ [0.2cm]
    1 & Random Baseline & 27.5 & 8.3 & \small learn tips on how to play the bass drum beat variation on the guitar in this free video clip on music theory and guitar lesson . \\
    \bottomrule
  \end{tabular}
  \caption{Example outputs of ground-truth text-and-video with hierarchical attention (8), text-only with ground-truth (5a), text-only with ASR output (5c), ASR output text-andv-video with hierarchical attention (9), action features with RNN (7) and action features only (6) models compared with the reference, the topic-based next neighbor (2b) and random baseline (1). Arranged in the order of best to worst summary in this table.}
  \label{tab:output_examples}
\end{table*}

\begin{figure*}[!h]
  \centering
  \renewcommand{\thefigure}{A1}
  \includegraphics[width=\linewidth]{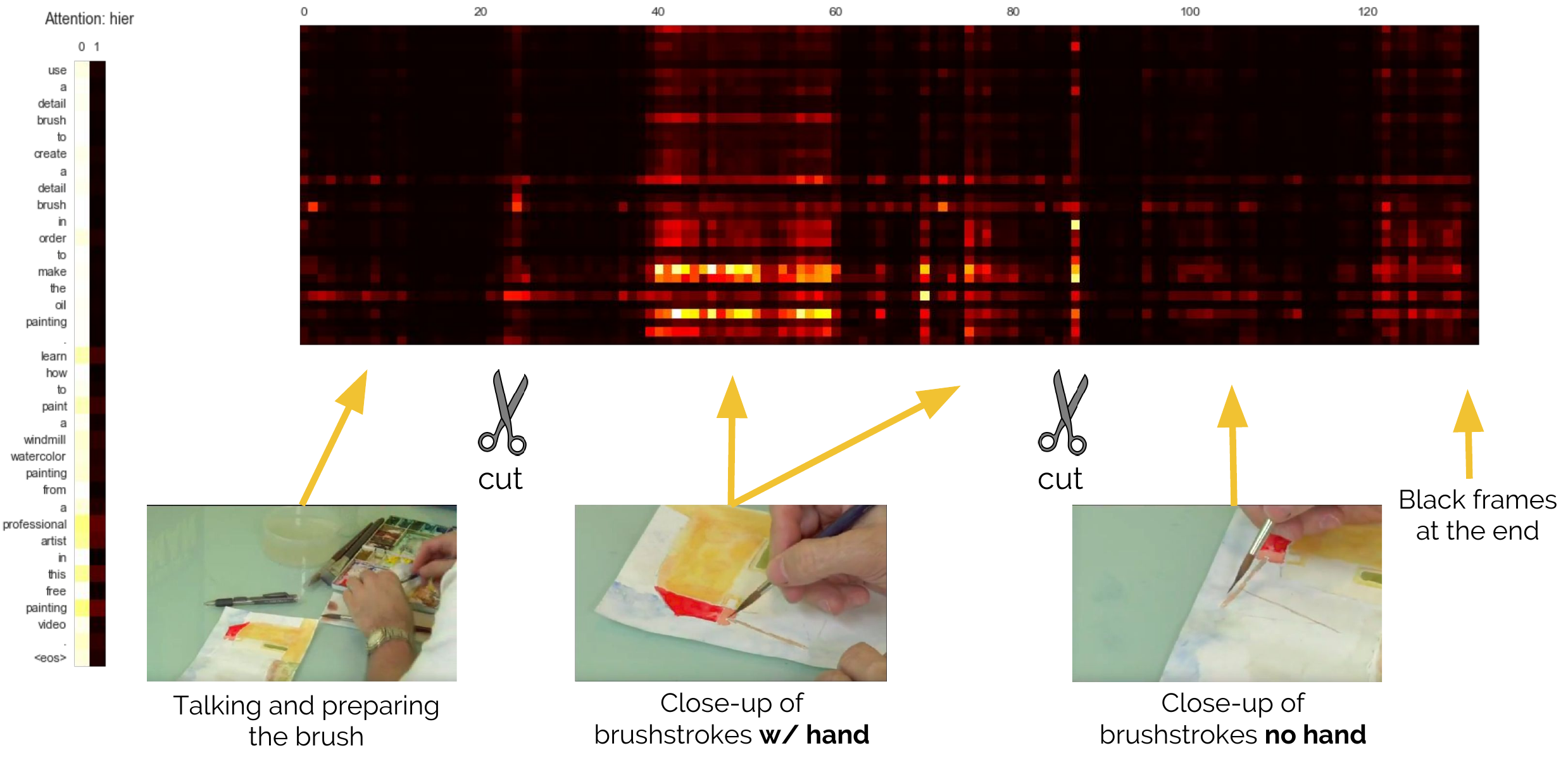}
  \caption{Visualizing Attention over Video Features.}
  \label{fig:attn}
\end{figure*}

\subsection{Output Examples from Different Models}
\label{ssec:supp_outputs}
Table~\ref{tab:output_examples} shows example outputs from our different
text-only and text-and-video models. The text-only model produces a fluent
output which is close to the reference.  The action features with the RNN
model, which sees no text in the input, produces an in-domain (``fly tying'''
and ``fishing'') abstractive summary that involves more details like
``equipment'' which is missing from the text-based models but is relevant. The
action features without RNN model belongs to the relevant domain but contains
fewer details. The nearest neighbor model is related to ``knot tying'' but not
related to ``fishing''. The scores for each of these models reflect their
respective properties. The random baseline output shows the output of sampling
from the random language model based baseline. Although it is a fluent output,
the content is incorrect. Observing other outputs of the model we noticed that
although predictions were usually fluent leading to high scores, there is scope
to improve them by predicting all details from the ground truth summary, like
the subtle selling point phrases, or by using the visual features in a
different adaptation model.

\subsection{Attention Analysis}
Figure~\ref{fig:attn} shows an analysis of the attention distributions using
the hierarchical attention model in an example video of painting. The vertical
axis denotes the output summary of the model, and the horizontal axis denotes
the input time-steps (from the transcript). We observe less attention in the
first part of the video where the speaker is introducing the task and preparing
the brush. In the middle half, the camera focuses on the close-up of brush
strokes with hand, to which the model pays higher attention over consecutive
frames. Towards the end, the close up does not contain the hand but only the
paper and brush, where the model again pays less attention which could be due
to unrecognized actions in the close-up. There are black frames in the very end
of the video where the model learns not to pay any attention. In the middle of
the video, there are two places with a cut in the video when the
camera shifts angle. The model has learned to identify these areas and uses it
effectively. From this particular example, we see the model using both
modalities very effectively in this task of the
summarization of open-domain videos.

\subsection{Human Evaluation Details}
\label{ssec:supp_human}

To understand the outputs generated for this task better, we ask workers on Amazon Mechanical Turk to compare outputs of unimodal and multimodal models with the ground-truth summary and assign a score between 1 (lowest) and 5 (highest) for four metrics: informativeness, relevance, coherence and fluency of generated summary. The annotators were shown the ground-truth summary and a candidate summary (without knowledge of the type of modality used to generate it). Each example was annotated by three workers. Annotation was restricted to English speaking countries. 129 annotators participated in this task.

\end{document}